\documentclass[runningheads]{llncs}
\usepackage[T1]{fontenc}
\usepackage{graphicx}
\usepackage{amssymb}   

\usepackage{hyperref} 

\usepackage{color}

\usepackage[english]{babel}
\usepackage{amsmath}
\usepackage{comment}
\usepackage{algorithm}
\usepackage{algpseudocode}

\begin{document}
\title{DTOC: Dynamic Tool Output Compression\\ 
for Adaptive Context Management in AI Agents}

\titlerunning{DTOC: Dynamic Tool Output Compression}
%
\author{Abhay Chaturvedi\inst{1} \and
Shreya Bhattacharya\inst{1}\orcidID{0009-0006-7889-1474} \and
Rashmika Gopalkrishnan\inst{1} \and
Peter van der Putten\inst{2,3}\orcidID{0000-0002-6507-6896}
}
\authorrunning{A. Chaturvedi et al.}
%
\institute{Product R\&D, Pegasystems, Bangalore, India\\
\email{\{abhay.chaturvedi,shreya.Bhattacharya,rashmika.gopalakrishnan\}@pega.com}
\and AI Lab, Pegasystems, Amsterdam, the Netherlands\\
\email{peter.vanderputten@pega.com}\\
\and LIACS, Leiden University, Leiden, the Netherlands\\
\email{p.w.h.van.der.putten@liacs.leidenuniv.nl}
}
\maketitle              
\begin{abstract}
As agent capabilities have grown, practical limitations increasingly stem from constrained context windows rather than model capacity. Common strategies, such as truncation, heuristic aging, and lossy summarization, may discard useful information or introduce hallucination risk. To address these challenges, we propose Dynamic Tool Output Compression (DTOC), a framework for scalable context management in LLM-based agents that models context updates as explicit and reversible operations within the agent reasoning loop. DTOC retains full tool outputs in external memory while inserting compact placeholders into the active context, enabling selective reconstruction when needed.
We formalize the DTOC mechanism, integrate it into a ReAct-style agent architecture, and provide a production-oriented implementation supporting on-demand restoration of compressed outputs. 
Experiments on DeepSWE reveal model-dependent effects: for responsive models (Sonnet 4.6, GPT-5.4), DTOC reduces input tokens (10.3 and 12.7\%) and agent steps (2.4 and 32.3\%), while increasing solve rates (2.5 and 1.5 times higher) and lowering cost per solved task (3 and 3.5 times lower cost per solved task). For the other models results are more mixed, with GPT-5.5 doubling solve rate and halving cost, but no impact on solve rate and negative impact on cost for the other models.
Ablation results show reversibility is critical: disable-only compression variants degraded performance, while full DTOC recovered baseline accuracy at substantially lower context cost. 
These findings indicate that explicit, reversible context management can improve the efficiency of long-horizon agent reasoning without degrading task performance.


\end{abstract}

\section{Introduction}

The primary challenge in maintaining long-horizon agentic trajectories is the inherent attention scarcity within the transformer architecture \cite{zhang2025memory,wan-etal-2026-compass,team2026longcat,anthropic2025context,yu2023trams}. Modern large language models (LLMs) rely on self-attention, where every token attends to every other token, resulting in $O(n^2)$ pairwise interactions as the context length 
$n$ grows \cite{vaswani2017attention}. As $n$ expands, the model’s ability to retrieve specific `needles’ from an increasingly dense informational `haystack’ degrades, reducing its capacity to surface earlier but task-critical details \cite{liu2024lost}. 
This limitation manifests empirically as `context rot', a degradation in the model’s effective reasoning horizon: although models may support large nominal context windows (e.g., 128k tokens), performance often declines well before these limits are reached \cite{liu2024lost}. 

These constraints become particularly acute in long horizon tasks, such as scientific and software engineering workflows, where (semi-)autonomous agents must integrate heterogeneous tool outputs over many steps \cite{yu2023trams,tziola2023autonomous,tallam2025autonomousagentsintegratedsystems,plaat2026agenticllmsurvey}. Over long trajectories, the accumulation of prompts, tool outputs, and intermediate reasoning saturates the available context, forcing truncation of earlier turns. Critically, the earliest messages often encode the user’s primary objectives, safety constraints, or key assumptions, leading subsequent reasoning to proceed on a progressively distorted representation of the original task \cite{lindenbauer2025gitgoodbench}. 

Existing context-management strategies often prove insufficient in agentic settings \cite{zhang2025memory}.
Hard truncation, in which older segments are removed once a token budget is exceeded, results in irreversible information loss and effectively forces planning under partial observability \cite{elastic2025context,anthropic2025context}. 
LLM-based summarization, used in systems such as OpenHands \cite{wang2025openhands} and Cursor \setcounter{footnote}{0}\footnote{\url{https://www.openhands.dev/}, \url{https://cursor.com/}}, mitigates token growth but can introduce abstraction errors, hallucinations, and trajectory elongation, causing agents to overlook failure signals and repeat unproductive actions \cite{lindenbauer2025gitgoodbench,jetbrains2025context}.
Other lossy approaches, such as Caveman-style compression\footnote{\url{https://github.com/juliusbrussee/caveman}}, face similar fidelity trade-offs, while static observation masking relies on recency or size heuristics and cannot easily recover relevant information from earlier reasoning steps \cite{lindenbauer2025gitgoodbench}.

To address these limitations, we introduce Dynamic Tool-Output Compression (DTOC), a framework that treats context management as an explicit agent action rather than a fixed system-level mechanism. Tool outputs are stored in an external memory and replaced in the active context with lightweight references, allowing agents to selectively hide and later restore information as needed. DTOC therefore provides reversible, low-risk context compression that preserves recoverability while reducing active-context growth during long-horizon reasoning.
A reference implementation built on OpenCode and used for the main experiments is available at \url{https://github.com/chaturvediabhay24/opencode}.

\section{Related Work}

The conceptual lineage of DTOC traces back to the ReAct framework 
\cite{yao2023reactsynergizingreasoningacting}, which showed that interleaving explicit reasoning traces (`thoughts’) with actions and observations enables more robust problem solving than reasoning alone \cite{yao2023reactsynergizingreasoningacting}. In ReAct, reasoning traces function as a form of inner speech that allows the model to induce, track, and revise action plans while integrating new evidence from the environment. However, ReAct was developed under the assumption of relatively short interaction histories and does not explicitly address the constraints imposed by finite context windows in long-horizon trajectories  \cite{yao2023reactsynergizingreasoningacting}.

Subsequent work on prompt compression, notably LLMLingua \cite{jiang2023llmlingua}, sought to address context limitations by using smaller language models to identify and remove low-importance tokens based on perplexity. LLMLingua reports substantial compression (up to 20× in some settings) for static prompts while maintaining task performance across multiple benchmarks \cite{jiang2023llmlingua}. 
In agentic settings, however, perplexity-based compression can be overly aggressive: tokens that appear linguistically predictable may carry high task-specific semantic value, such as unique identifiers, numeric parameters, or error codes in execution logs. Moreover, LLMLingua operates as a fundamentally lossy, one-way transformation, discarding information that cannot be fully reconstructed from the compressed prompt \cite{jiang2023llmlingua}. DTOC instead targets the dominant source of context growth, tool outputs, and controls their visibility through a non-destructive mechanism that hides or reveals content without irreversible deletion.

In parallel, a broader shift toward `context engineering' has emerged as a formal discipline concerned with the systematic optimization of information payloads presented to LLMs \cite{mei2025survey,hua2025context,zhang2025agentic}.
Anthropic’s \cite{anthropic2025context} work on `Agent Skills’ and `Progressive Disclosure' exemplifies this direction: agents maintain lightweight references or handles to external resources and selectively load full content only when necessary, rather than streaming all information into the live context at once \cite{anthropic2025context}. DTOC generalizes this principle from individual files or queries to the entire conversational state, treating the evolving interaction history as a dynamic repository that the agent can explicitly load or unload as its reasoning unfolds, enabling agent control over system heuristics.

Closely related is Stanford’s ACE (Agentic Context Engineering)framework \cite{zhang2025agentic}, which conceptualizes contexts as `evolving playbooks' that accumulate, refine, and organize strategies through a generator–reflector–curator cycle \cite{zhang2025agentic}. ACE focuses on high-level adaptation of strategy representations and demonstrates how naive, end-to-end rewriting of long contexts can induce `context collapse', where crucial details are inadvertently erased during iterative summarization \cite{zhang2025agentic}. DTOC is complementary to ACE: whereas ACE manages the structure and evolution of strategic playbooks, DTOC supplies the low-level token management and tool-output governance needed to keep these playbooks, and the evidentiary traces they depend on, within the model’s effective attention span.

\section{Methodology: Dynamic Tool Output Compression}

This section formalizes Dynamic Tool Output Compression (DTOC) as a structured, non-destructive context management framework. We describe the general mechanism (Section~\ref{sec:method-general}), detail the implementation via an explicit tool-based interface (Section~\ref{sec:method-implementation}), specify the underlying data structures (Section~\ref{sec:method-datastructures}), and finally situate DTOC within the broader context engineering landscape (Sections~\ref{sec:contextengineeing}, \ref{sec:ace-playbook}).

\subsection{General Mechanism Overview}
\label{sec:method-general}

DTOC formulates context management as an explicit agent-level operation over a persistent external memory of tool outputs. At each step, the agent maintains a minimal active context comprising the current user query, its reasoning trace, and a curated subset of prior tool outputs. All outputs are stored in a structured repository indexed by lightweight identifiers, aligning with `Agent Skills' and `Progressive Disclosure' paradigms where agents reference resources via handles rather than raw content \cite{anthropic2025context}.

Instead of compressing or discarding information, DTOC enables explicit retrieval and unloading actions, allowing the agent to dynamically manage context under token constraints. This yields a non-destructive visibility layer in which outputs remain persistently stored and recoverable, with access governed by the agent’s learned retrieval policy rather than irreversible truncation. Consequently, DTOC shifts the bottleneck from storage capacity to efficient access and retrieval.

\subsection{Implementation via Explicit Tool-Based Control}
\label{sec:method-implementation}

\begin{figure}[tbp]
\centering
\includegraphics[
    width=\textwidth,
    trim=0cm 6cm 4cm 0cm, 
    clip
]{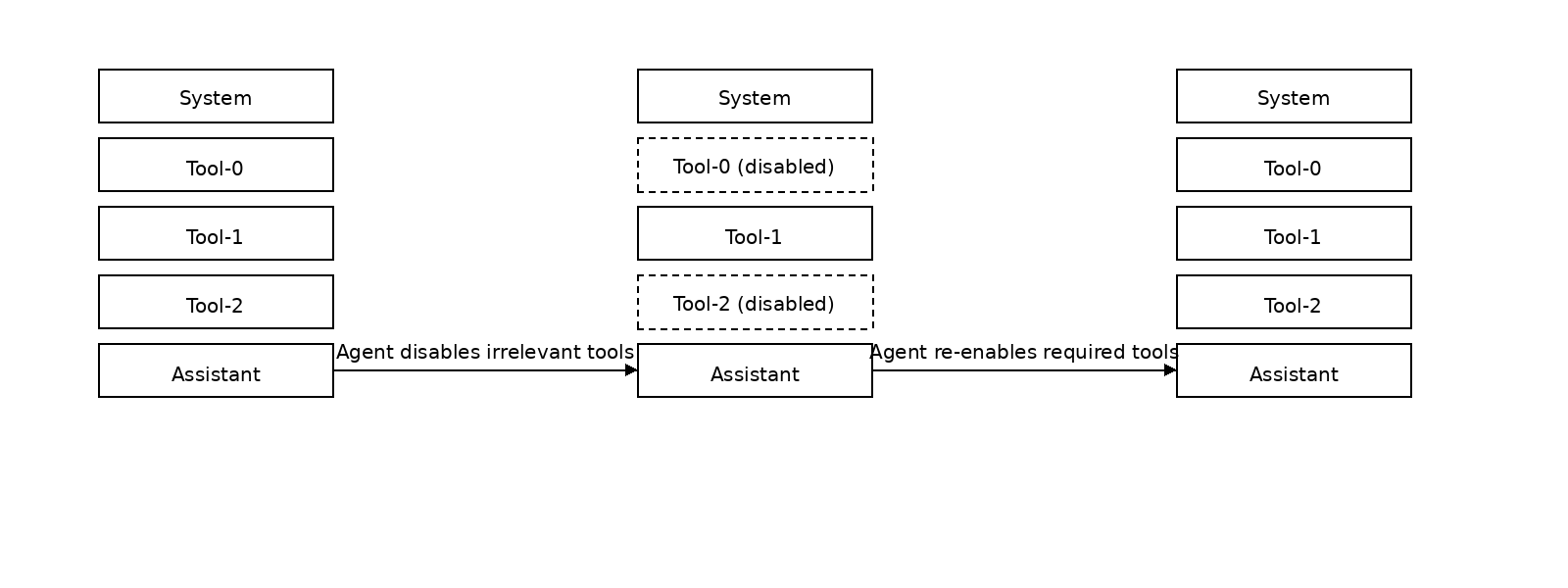}
\caption{Architecture of DTOC. The figure shows reversible, agent-controlled context compression. Initially, all tool outputs are included in the active context (left). The agent selectively disables irrelevant tool outputs to reduce token usage while retaining their metadata (middle). When needed, previously disabled outputs can be re-enabled without loss of information (right), enabling progressive disclosure and fine-grained context engineering under orchestrator-level control.}
\label{fig:arch}
\end{figure}

In practice, DTOC is realized through a dedicated context management tool that exposes visibility control as an explicit agent action (Figure~\ref{fig:arch}). Rather than relying on provider-specific tool identifiers (e.g., OpenAI or Anthropic tool IDs), the orchestration layer assigns custom identifiers at execution time and wraps each tool response in a structured JSON object containing three fields: \texttt{tool\textunderscore key}, a unique identifier; \texttt{tool\textunderscore result}, the output content; and \texttt{estimated\textunderscore tokens}, an approximate token count used for context budgeting.

The agent interacts with a \texttt{manage\textunderscore context} tool parameterized by two lists, \texttt{enable} and \texttt{disable}, each specifying a set of tool keys. Invoking this tool updates output visibility within the active context. Disabled outputs are replaced with compact metadata-only placeholders containing the tool key, timestamp, and token estimate, while enabled outputs restore their full \texttt{tool\textunderscore result} from the orchestrator. This mechanism decouples visibility management from provider APIs, yielding a model-agnostic and unified representation of visible, hidden, and recoverable outputs.

Trigger conditions for context compression are conveyed through a natural language tool description (e.g., invoking the tool when the context nears its limit and certain outputs become irrelevant). This design delegates compression policy to the agent rather than fixed heuristics, aligning with DTOC’s agent-centric philosophy. By exposing token estimates for each output, the agent can make informed trade-offs between retaining useful information and maintaining a compact context.

\subsection{DTOC Data Structures and Tool Interface} 
\label{sec:method-datastructures}

To make the DTOC mechanism concrete, we formally specify the key data structures and tool interfaces used in our implementation. Algorithm~\ref{alg:dtoc} presents the core orchestrator logic, including the \textsc{ToolOutput} record structure, the persistent store, the context assembly procedure, and the \texttt{manage\_context} interface.

\begin{algorithm}[tb]
\caption{DTOC Orchestrator: Data Structures and Context Management}\label{alg:dtoc}
\begin{algorithmic}
\Statex \textbf{Record} \textsc{ToolOutput}:
\Statex \quad \texttt{tool\_key}: String \Comment{Orchestrator-generated unique ID, e.g.\ \texttt{"tool\_001"}}
\Statex \quad \texttt{tool\_result}: String \Comment{Raw content returned by external tool}
\Statex \quad \texttt{estimated\_tokens}: Integer \Comment{Heuristic token-count approximation}
\Statex \quad \texttt{timestamp}: DateTime \Comment{Invocation time}
\Statex \quad \texttt{visible}: Boolean \Comment{Indicates inclusion in the active context}
\Statex
\Statex \textbf{State:} Persistent store $\mathcal{S}$: \textsc{ToolKey} $\rightarrow$ \textsc{ToolOutput}
\Statex
\Procedure{AssembleContext}{$\mathcal{S}$}
    \State $\mathcal{C} \gets \emptyset$
    \For{each $(k, o) \in \mathcal{S}$}
        \If{$o.\texttt{visible} = \textbf{true}$}
            \State $\mathcal{C} \gets \mathcal{C} \cup \{o\}$ \Comment{Include full tool output}
        \Else
            \State $\mathcal{C} \gets \mathcal{C} \cup \{\textsc{Placeholder}(k, o.\texttt{estimated\_tokens}, o.\texttt{timestamp})\}$
        \EndIf
    \EndFor
    \State \Return $\mathcal{C}$
\EndProcedure
\Statex
\Procedure{ManageContext}{enable: \texttt{List[ToolKey]}, disable: \texttt{List[ToolKey]}}
    \For{each $k \in \texttt{disable}$}
        \State $\mathcal{S}[k].\texttt{visible} \gets \textbf{false}$
    \EndFor
    \For{each $k \in \texttt{enable}$}
        \State $\mathcal{S}[k].\texttt{visible} \gets \textbf{true}$
    \EndFor
    \State \Return \textsc{StatusReport}(\textit{modified keys}, $\Delta$~\textit{active tokens})
\EndProcedure
\end{algorithmic}
\end{algorithm}

Each \textsc{ToolOutput} record associates a unique orchestrator-assigned identifier (\texttt{tool\_key}) with the raw content returned by an external tool (\texttt{tool\_result}), along with a heuristic estimate of its token footprint (\texttt{estimated\_tokens}), an invocation timestamp, and a \texttt{visible} flag that determines whether the content is included in the active context. The orchestrator maintains a persistent store $\mathcal{S}$ mapping each \textsc{ToolKey} to its corresponding record.

At each reasoning step, the \textsc{AssembleContext} procedure constructs the active context $\mathcal{C}$. Outputs marked as visible are incorporated in full, whereas hidden outputs are replaced by compact placeholders that retain only essential metadata, namely the \textsc{ToolKey}, estimated token count, timestamp, and an implicit hidden-status indicator. This design preserves referential integrity while minimizing token consumption.

Interaction with the context management mechanism is mediated through the \textsc{ManageContext} procedure, which accepts two sets of \textsc{ToolKey} values corresponding to enable and disable operations, respectively. The procedure updates the visibility flags in $\mathcal{S}$ and returns a \textsc{StatusReport} summarizing the modified entries and the resulting change in active token count. This immediate feedback enables the agent to reason explicitly about the trade-off between context breadth and token efficiency.

The interface design satisfies three key properties. First, all compression operations are reversible, as the complete \texttt{tool\_result} is persisted in $\mathcal{S}$ irrespective of its visibility status. Second, the agent has explicit access to metadata such as token estimates and timestamps, supporting informed and adaptive context selection. Third, the mechanism is provider-agnostic, relying solely on orchestrator-level abstractions rather than vendor-specific tool-calling semantics.

We estimate token counts with a character-count heuristic (\texttt{len(content)/4}) applied when the orchestrator wraps each tool output. The \texttt{manage\_context} tool description provided verbatim to all models reads: \emph{``Use this tool when you feel the context limit is increasing and there are some tool outputs whose content is no longer relevant to the current reasoning step. One can disable such outputs to free up context space, and re-enable them later if needed.''} No explicit trigger threshold is imposed; compression timing is delegated entirely to the agent. Placeholders for disabled outputs consume approximately 30--50 tokens regardless of original output size. When the active context approaches the model's native limit, the underlying provider applies its own truncation; DTOC aims to prevent reaching this threshold through proactive agent-driven compression but does not override provider-level behavior.

\subsection{Context Engineering Perspective}
\label{sec:contextengineeing}

Our methodology adopts the emerging lens of context engineering, formalizing the interaction history and tool-output store as an information payload to be actively optimized rather than passively accumulated \cite{mei2025survey}. 
Concretely, DTOC operationalizes context engineering \cite{mei2025survey,hua2025context,zhang2025agentic} at two levels.
First, it generalizes the `Progressive Disclosure' principle from individual resources (e.g., files or queries) to the full conversational trajectory: past messages and tool outputs remain available for selective re-instantiation into the live context, but are not maintained there by default \cite{anthropic2025context}. This shifts context management from static inclusion to dynamic, demand-driven access.
Second, DTOC introduces a compact action vocabulary, conceptually including operations such as select, enable, disable, and retrieve, that the agent can invoke to manage its memory footprint. This elevates context selection into a learnable component of the policy, rather than a fixed system heuristic.

\subsection{Relation to ACE and Playbook Evolution}
\label{sec:ace-playbook}

Finally, we situate DTOC within the Agentic Context Engineering (ACE) framework \cite{zhang2025agentic}, which views contexts as `evolving playbooks' that accumulate and refine strategies via a generator–reflector–curator cycle. In our experimental setup, high-level strategy adaptation (e.g., switching debugging tactics or adjusting research plans) corresponds to ACE’s notion of playbook evolution, while DTOC provides the low-level token and tool-output management required to keep these playbooks within the model’s effective attention span. 

This separation of concerns enables us to disentangle whether strategic failures arise from genuine reasoning errors or from context pathologies such as context collapse, where critical details are lost due to repeated summarization or truncation \cite{zhang2025agentic}. By instrumenting both high-level decision traces and the underlying DTOC operations, our methodology quantifies the extent to which dynamic, non-destructive control over tool-output visibility improves success rates and sample efficiency in long-horizon agentic tasks.

\section{Experimental Setup}

We evaluate DTOC on six tasks from DeepSWE \cite{datacurve2026deepswe}, a benchmark of real-world open-source software engineering tasks, using five frontier models under two conditions: DTOC ON and DTOC OFF (Table \ref{tab1}). We use DeepSWE \cite{datacurve2026deepswe} because all tasks require implementing complete, non-trivial features that produce extended agent trajectories (30--285 steps), providing a naturalistic stress test for context management under realistic token accumulation. 
The five models were selected to span three axes of variation that are directly relevant to context management. First, provider diversity: we include Anthropic (Sonnet, Opus), OpenAI (GPT‑5.4, GPT‑5.5), and Google (Gemini) to ensure our findings are not idiosyncratic to a single architecture or training methodology. Second, context window size: the selected models cover native capacities of 128K (GPT‑5.4/GPT‑5.5), 200K (Sonnet/Opus), and 1M tokens (Gemini Flash), allowing us to test whether DTOC’s benefit attenuates as native context capacity grows. Third, capability tier and reasoning style: by pairing a mid‑tier and a high‑end model from each provider (Sonnet vs. Opus; GPT‑5.4 vs. GPT‑5.5) we evaluate whether DTOC benefits scale with model capability, while Gemini Flash exemplifies a “large‑context, fast‑inference” design point. Together, these choices let us isolate effects of provider, native context size, and model capability on DTOC’s effectiveness. 

\begin{table}[tbp]
\centering
\caption{Experimental Grid}
\label{tab1}
\begin{tabular}{l p{10cm}}
\textbf{Component} & \textbf{Setting} \\
\hline
Benchmark &
DeepSWE (6 real-world open-source SWE tasks; Python and TypeScript) \\

Models &
Claude Sonnet 4.6 (200K); Claude Opus 4.8 (200K); GPT-5.4 (128K); GPT-5.5 (128K); Gemini 3.5 Flash (1M) \\

Conditions &
DTOC ON (agent has \texttt{manage\_context} tool); DTOC OFF (no compression, full context retained) \\

Metrics &
Task solve rate (reward), input tokens, agent steps, wall-clock duration, cost per solved task \\
Agent &
opencode v0.0.0-dtoc-implementation (ReAct-style, shared prompts, identical tool configurations)
\end{tabular}
\end{table}

\subsection{Tasks and Environments}
\label{tasks}

DeepSWE requires implementing complete features capable of passing each pro\-ject's full test suite. The tasks span Python projects (Narwhals, FastAPI, sqlite-utils, and Mashumaro) and TypeScript projects (drizzle-orm and ts-pattern), ranging from the implementation of a single combinator (\texttt{matchEach}) to multi-file rolling-window features. These tasks generate long agent trajectories (30--285 steps) with substantial tool-output accumulation, making DeepSWE a natural stress test for context management.

Agents are executed using OpenCode, a ReAct-style coding agent that interleaves reasoning and tool invocation. The available toolset includes file operations (\texttt{read}, \texttt{write}, \texttt{edit}, and \texttt{apply\textunderscore patch}), code-intelligence tools (\texttt{glob}, \texttt{grep}, and \texttt{lsp}), and shell execution via \texttt{bash}.
Each model--condition pair was run on all assigned tasks using identical agent configurations, with tool outputs wrapped in structured JSON envelopes containing orchestrator-assigned identifiers, raw results, and token estimates. In the DTOC-ON condition, the agent is additionally provided with the \texttt{manage\textunderscore context} tool, which enables dynamic visibility control over previously generated tool outputs (Section~\ref{sec:method-implementation}). All other tools, prompts, and configuration settings are held constant across experimental conditions.
Task success is measured as a binary outcome. A task receives a reward of $1.0$ if all project tests pass successfully and $0.0$ otherwise.

\section{Results} 

We evaluate DTOC on the DeepSWE \cite{datacurve2026deepswe} benchmark as per the set up described in the previous section, along with a small-scale ablation study on proprietary data.

\subsection{DeepSWE Benchmark}

Table~\ref{tab:deepswe} summarizes the aggregate DeepSWE results per model and Table~\ref{tab:deepswe-tasks} shows the solve matrix by task and model, under DTOC ON and OFF conditions. These  results reveal model-family-dependent compression behaviors. DTOC's effectiveness depends on each model's native reasoning style, verbosity, and implicit context-management strategies.

\begin{table}[tbp]
\centering
\caption{DeepSWE benchmark: aggregate results per model (6 real-world SWE tasks). $\Delta$ columns show relative change with DTOC ON vs.\ OFF; negative values indicate improvement. SR is Solve Rate, CS is Cost Solved. Denominators below 6 reflect tasks excluded due to infrastructure timeouts.\protect\footnotemark}
\label{tab:deepswe}
\begin{tabular}{|l|c|c|c|c|c|c|}
\hline
\textbf{Model} &
\textbf{SR} &
\textbf{SR} &
\textbf{Input} &
\textbf{Steps $\Delta$} &
\textbf{CS} &
\textbf{CS} \\
 &
\textbf{ON} &
\textbf{OFF} &
 \textbf{Tokens $\Delta$} &
 &
\textbf{ON} &
\textbf{OFF} \\
\hline
Claude Sonnet 4.6  & 3/6 & 1/5  & $-10.3\%$ & $-2.4\%$  & \$8.65  & \$26.27  \\
\hline
GPT-5.4            & 3/6  & 2/6  & $-12.7\%$ & $-32.3\%$ & \$2.64  & \$9.30   \\
\hline
GPT-5.5            & 2/6  & 1/6  & $-0.6\%$  & $+14.3\%$ & \$49.93 & \$100.49 \\
\hline
Claude Opus 4.8    & 3/6  & 3/6  & $+18.4\%$ & $+8.2\%$  & \$21.50 & \$19.64  \\
\hline
Gemini 3.5 Flash   & 1/5  & 1/5  & $+48.7\%$ & $+30.2\%$ & \$18.11 & \$13.28  \\
\hline
\end{tabular}
\end{table}
\footnotetext{Mashumaro was not evaluated for Sonnet OFF and Gemini (both conditions) due to infrastructure timeouts; denominators reflect completed runs only.}

DTOC has a strongly positive effect on Claude Sonnet 4.6. The solve rate increases from $1/5$ ($20\%$) to $3/6$ ($50\%$). Average input tokens decrease from $369{,}666$ to $331{,}615$ ($-10.3\%$), output tokens decrease from $104{,}442$ to $93{,}804$ ($-10.2\%$), and mean task duration decreases by $106\,\mathrm{s}$ (from $2{,}318\,\mathrm{s}$ to $2{,}212\,\mathrm{s}$). At the task level, \textit{ts-pattern} exhibits a $3.6\times$ speedup ($32$ versus $117$ steps), \textit{sqlite-utils} succeeds with DTOC but fails almost immediately without it ($2$ steps), and \textit{drizzle-orm} is fully solved only in the DTOC condition. Cost per solved task decreases by approximately $67\%$, from $\$26.27$ to $\$8.65$.

GPT-5.4 is highly compression-responsive and achieves the best overall efficiency. DTOC reduces input tokens by $12.7\%$ (from $129{,}824$ to $113{,}374$), cache-read tokens by $26.8\%$, and agent steps by $32.3\%$ (from $54.2$ to $36.7$). The solve rate improves from $33\%$ to $50\%$, while average duration decreases by $59\,\mathrm{s}$. Notably, \textit{mashumaro} is solved with DTOC ($72/72$ tests passing in $33$ steps) but fails without DTOC ($67/72$ tests passing in $73$ steps), suggesting that DTOC preserves critical information during long trajectories. GPT-5.4 combined with DTOC achieves the lowest cost per solved task, reducing cost from \$9.30 to \$2.64.

GPT-5.5 exhibits a modest but positive response to DTOC. The solve rate doubles from $1/6$ to $2/6$, and mean duration decreases from $790\,\mathrm{s}$ to $740\,\mathrm{s}$, making GPT-5.5 the fastest model overall at approximately $12.3$ minutes per task. Input token savings are negligible ($-0.6\%$), while agent steps increase by $14.3\%$, suggesting that frequent \texttt{manage\textunderscore context} invocations generate overhead without substantially reducing context size. DTOC enables a complete solution for \textit{ts-pattern} and yields substantially greater partial progress on \textit{fastapi} ($519$ tests passed versus $43$ without DTOC). Cost per solved task decreases from \$100.49 to \$49.93.

Claude Opus 4.8 exhibits task-dependent trade-offs with no aggregate improvement in solve rate, remaining at $3/6$ in both conditions. Under DTOC, input tokens increase by $18.4\%$, agent steps increase by $8.2\%$, and execution time increases by $5.7\%$. However, outcomes vary across tasks. For example, \textit{mashumaro} succeeds only with DTOC, whereas \textit{sqlite-utils} succeeds only without DTOC. These results suggest that certain compression decisions may occasionally remove context that remains relevant. Because Opus generates substantially more output tokens than the other models, the relative benefits of DTOC are reduced and partially offset by the overhead of context-management operations.

DTOC has a negative effect on Gemini 3.5 Flash. Input tokens increase by $48.7\%$, agent steps increase by $30.2\%$, and execution time increases by $35.8\%$, while the solve rate remains unchanged at $1/5$. Cost per solved task increases by approximately $36\%$. Inspection of execution traces suggests that Gemini frequently performs small compression operations that introduce overhead without producing meaningful reductions in active context size. Its incremental action style generates many short outputs and tool interactions, causing management overhead to outpace compression gains. These findings suggest that model-specific compression policies may be necessary to fully realize DTOC's benefits.
\begin{table}[tbp]
\centering
\caption{DeepSWE per-task solve matrix ($\checkmark$ = solved, $\times$ = not solved, \textemdash{} = not evaluated).}
\label{tab:deepswe-tasks}
\begin{tabular}{l|cc|cc|cc|cc|cc}
\hline
& \multicolumn{2}{c|}{\textbf{Sonnet 4.6}}
& \multicolumn{2}{c|}{\textbf{GPT-5.4}}
& \multicolumn{2}{c|}{\textbf{GPT-5.5}}
& \multicolumn{2}{c|}{\textbf{Opus 4.8}}
& \multicolumn{2}{c}{\textbf{Gemini 3.5}} \\
\textbf{Task}
& ON & OFF
& ON & OFF
& ON & OFF
& ON & OFF
& ON & OFF \\
\hline
narwhals (Python)
& $\times$ & $\times$
& $\times$ & $\times$
& $\times$ & $\times$
& $\times$ & $\times$
& $\times$ & $\times$ \\

drizzle-orm (TS)
& $\checkmark$ & $\times$
& $\times$ & $\checkmark$
& $\checkmark$ & $\checkmark$
& $\checkmark$ & $\checkmark$
& $\times$ & $\times$ \\

fastapi (Python)
& $\times$ & $\times$
& $\times$ & $\times$
& $\times$ & $\times$
& $\times$ & $\times$
& $\times$ & $\times$ \\

sqlite-utils (Python)
& $\checkmark$ & $\times$
& $\checkmark$ & $\times$
& $\times$ & $\times$
& $\times$ & $\checkmark$
& $\times$ & $\times$ \\

ts-pattern (TS)
& $\checkmark$ & $\checkmark$
& $\checkmark$ & $\checkmark$
& $\checkmark$ & $\times$
& $\checkmark$ & $\checkmark$
& $\checkmark$ & $\checkmark$ \\

mashumaro (Python)
& $\times$ & \multicolumn{1}{c|}{\textemdash}
& $\checkmark$ & $\times$
& $\times$ & $\times$
& $\checkmark$ & $\times$
& \multicolumn{1}{c}{\textemdash} & \multicolumn{1}{c}{\textemdash} \\
\hline

\textbf{Total}
& \textbf{3/6} & \textbf{1/5}
& \textbf{3/6} & \textbf{2/6}
& \textbf{2/6} & \textbf{1/6}
& \textbf{3/6} & \textbf{3/6}
& \textbf{1/5} & \textbf{1/5} \\
\hline
\end{tabular}%
\end{table}

Task difficulty as shown in per-task solve matrix Table~\ref{tab:deepswe-tasks} also moderates DTOC's effectiveness. Two tasks, \textit{narwhals} and \textit{fastapi}, remain unsolved by all evaluated models, indicating intrinsic difficulty beyond context-management considerations. In contrast, \textit{ts-pattern} is solved in $9$ of $10$ model-condition pairs and serves as a near-ceiling baseline. DTOC's largest effects appear on medium-difficulty tasks. For example, \textit{sqlite-utils} and \textit{mashumaro} are solved only under DTOC for specific models, implying that context curation can be the decisive factor separating success from failure when tasks are challenging but still tractable.

\subsection{Ablation Study}

We did carry out a small-scale explorative ablation study on real-world proprietary data and code.
To isolate the contribution of individual DTOC components, we compare four configurations: no compression, tool tracking only, disable-only, and full enable+disable (Table \ref{tab5}). 

The results reveal that implementing tool tracking without the ability to disable context (`+ Tool tracking, no disable') resulted in a slight performance degradation to 77.8\% success rate while maintaining high context usage (87K tokens). The high frequency of file re-reads (4.2 average) suggests that tracking alone introduces significant overhead and necessitates a recovery mechanism to be effective.
Interestingly, the `Disable only' configuration significantly outperformed the uncompressed baseline, achieving an 81.4\% success rate with only 57K tokens. This indicates that aggressive context pruning can improve model focus by removing "noise," though the non-zero re-read average (0.1) highlights a persistent risk of irreversible information loss. The Full DTOC configuration achieved the optimal balance, reaching the highest overall success rate of 83.6\% using 61K tokens. By integrating both disabling and dynamic re-enabling capabilities, the framework eliminated file re-reads entirely (0.0). This confirms that the ability to recover previously compressed context is essential for achieving peak reliability, allowing the model to surpass the baseline performance while maintaining a significantly leaner context footprint.

\begin{table}[tbp]
\centering
\caption{Ablation of DTOC components}
\resizebox{\textwidth}{!}{%
\begin{tabular}{lllll}
\textbf{Configuration} &
\textbf{Success rate (\%)} &
\textbf{Context (tokens)} &
\textbf{Notes} &
\textbf{File re-reads (avg)} \\
\hline
No compression (baseline)      & 78.2 & 89K & —                     & 0.0 \\
+ Tool tracking, no disable    & 77.8 & 87K & Overhead only         & 4.2 \\
+ Disable only (no re-enable)  & 81.4 & 57K & Information loss risk & 0.1 \\
+ Full DTOC (enable + disable) & 83.6 & 61K& Best result           & 0.0
\end{tabular}}
\label{tab5}
\end{table}

\section{Discussion}

\subsection{Reflection on Results}

DeepSWE reveals clear model-family differences in DTOC effectiveness. Sonnet 4.6, GPT-5.4 and GPT-5.5  achieve higher solve rates and lower costs, while Opus 4.8 and Gemini 3.5 Flash show little or no aggregate benefit. For Opus, this neutrality masks complementary task-level effects: DTOC enables solving \textit{mashumaro}, whereas DTOC OFF solves \textit{sqlite-utils}, suggesting that DTOC reshapes rather than consistently expands the model's capability frontier.

Three model-level factors may explain the observed variation. First, native context-management behavior appears important: models that already avoid redundant tool interactions (e.g., GPT-5.4) amplify DTOC's benefits, whereas models that employ many small reasoning steps (e.g., Gemini) generate management overhead faster than DTOC can reduce context size. Second, output verbosity influences effectiveness; Opus produces substantially larger outputs (131K avg), reducing the relative savings achievable through compression. Third, instruction-following fidelity may play a role. Models that interpret the generic \textit{manage\_context} prompt as a signal to perform frequent shallow compressions appear to perform worse than models that apply compression selectively. These findings suggest that DTOC should be adapted to individual model families or guided by learned, model-specific policies. 

For medium-difficulty tasks such as \textit{sqlite-utils} and \textit{mashumaro}, DTOC is often the factor that separates success from failure. In contrast, easier tasks such as ts-pattern are solved consistently regardless of compression, while more difficult tasks such as narwhals and fastapi remain unsolved across all evaluated conditions. Sonnet's failure on sqlite-utils without DTOC, where execution stalled after only two steps, illustrates how context saturation can contribute to premature task abandonment. DTOC mitigates such failures by preserving a manageable active context throughout long-horizon trajectories.

For compression-responsive models, DTOC yields substantial economic benefits, reducing cost per solved task by approximately 67--72\% for Sonnet 4.6 (67\%) and GPT-5.4 (72\%). These gains arise from both direct token savings and improved solve rates, which amortize execution overhead across a larger number of successful outcomes. In contrast, models exhibiting increased token consumption under DTOC, such as Opus and Gemini, incur corresponding cost penalties, highlighting the need for model-specific deployment.

\subsection{Limitations}

Several limitations qualify these findings. 
First, the current DeepSWE task selection  contains only six feature-request tasks, limiting statistical power and domain coverage; refactoring and software architecture tasks were not evaluated. Expanding DeepSWE coverage to include additional task types and programming languages, such as Go, JavaScript, and Rust, would improve the breadth of the evaluation. 
Second, all experiments used a single generic \textit{manage\textunderscore context} prompt, which may not align equally well with the instruction-following behavior of all models. 
Third, the binary success metric, defined as whether all tests pass, obscures meaningful differences in partial progress. 
Finally, only a single trial was performed for each model--task--condition combination, leaving the influence of stochastic variability unmeasured. Additionally, the DeepSWE evaluation compares only DTOC ON versus OFF, without direct comparison against alternative context-management approaches such as LLMLingua~\cite{jiang2023llmlingua}, prompt truncation, or extractive summarization. Direct comparison is non-trivial because these methods target static prompts rather than dynamic multi-turn agent trajectories, but establishing relative performance remains an important direction. We note that DTOC is complementary to token-level compression: methods like LLMLingua could be applied \emph{within} individual tool outputs before DTOC manages their inter-output visibility.

\subsection{Future Work}

This paper should be seen as an introduction to the method and a proof of principle. Several directions merit further investigation to more robustly investigate its behavior in different contexts and expand into new feature sets and directions. 
First, larger benchmark suites, multiple independent runs and additional baseline comparisons  would enable statistical testing of observed effects. We experimented with other benchmarks as well, such as HotPotQA and proprietary company data and tasks, but we do recommend to expand DeepSWE coverage first, to keep the focus on long horizon tasks where context management matters most. 
Second, replicating ablation studies on DeepSWE could further isolate the impact of specific DTOC capabilities under different contexts. Initial experiments on proprietary use cases indicate that full DTOC outperforms DTOC that just turns context off versus no DTOC. 
Third, budget-aware prompting could expose real-time token counts and context-budget thresholds, helping agents make more informed compression decisions. Also, hybrid compression mechanisms that retain machine-generated summaries within placeholders may preserve useful preview information while still reducing context size, potentially benefiting models that prefer frequent incremental compression.
Finally, model-adaptive compression policies could be made visible through adjacent AI methods such as process mining, learned from trajectories or selected dynamically by the orchestrator to better align compression behavior with model-specific reasoning patterns.

\section{Conclusion}

Dynamic Tool Output Compression (DTOC) addresses context management as a first-class problem in agentic systems, rather than a secondary implementation detail. 
By allowing agents to explicitly control the visibility of past tool outputs, DTOC reframes compression from an irreversible, lossy operation into a reversible decision embedded in the agent’s reasoning loop.

Empirically, DTOC yields model-dependent effects. For compression-respon\-sive models (Sonnet 4.6, GPT-5.4, GPT-5.5), observed solve rates improve (e.g., from 20\% to 50\% for Sonnet, 33\% to 50\% for GPT-5.4) and cost per solved task decrease by 67--72\%. For Claude Opus 4.8 and Gemini 3.5 Flash, solve rates remain unchanged while token consumption increases by 18--49\%, indicating that DTOC should not be deployed without model-specific tuning. These findings are based on single runs per condition and require validation through repeated trials.

Qualitative analysis of agent trajectories reveals that agents spontaneously adopt a repertoire of compression strategies, such as breadth-then-depth exploration, iterative refinement, and chunking that are sensitive to task structure and uncertainty. This diversity supports the central hypothesis that compression decisions are best made by the agent itself, conditioned on its evolving beliefs and plan, rather than imposed via fixed external rules. DTOC thus forms a bridge between low-level context-window mechanics and high-level planning, allowing compression to be optimized in situ as part of the reasoning process.

Taken together, these results suggest that context engineering should be treated as a core dimension of agent design, on par with planning, tool use, and memory. DTOC provides a practical, model-agnostic framework for doing the same. It is simple to implement, requiring only orchestrator-level tracking of tool outputs and a single additional tool in the agent's action space, compatible with existing ReAct-style agents and tool ecosystems, and effective for models whose reasoning style aligns with selective, deliberate compression. The explicit tool-based interface, with orchestrator-assigned identifiers and structured JSON envelopes, ensures that DTOC can be deployed on any LLM platform without relying on provider-specific APIs or tool-call mechanisms. Looking forward, combining DTOC with learned compression policies, hierarchical summaries, and multi-agent coordination mechanisms offers a rich research agenda. As models continue to grow and tasks become more complex, we expect DTOC-like mechanisms to become a foundational component of scalable, production-grade agentic systems, which are lossless, agent-driven, and reversible.


\begin{credits}

\subsubsection{\discintname}
The authors have no competing interests to declare that are relevant to the content of this article.
\end{credits}

\bibliographystyle{splncs04}
\bibliography{dtoc}

\end{document}